\documentclass[runningheads]{llncs}
\usepackage[most]{tcolorbox}
\usepackage[T1]{fontenc}
\usepackage{graphicx}
\usepackage{float} 
\usepackage{xcolor} 
\usepackage{tabularx} 
\usepackage{listings} 
\usepackage{siunitx} 
\usepackage{booktabs}
\usepackage{array} 
\restylefloat{table}
\usepackage{caption} 
\usepackage{subcaption}

\begin{document}

\title{Enhancing Debunking Effectiveness through LLM-based Personality Adaptation}

\titlerunning{Enhancing Debunking Effectiveness through Personality Adaptation}
%
\author{
Pietro Dell'Oglio\inst{1}\thanks{Corresponding author}  \and
Alessandro Bondielli\inst{2} \and
Francesco Marcelloni\inst{1}\and
Lucia C. Passaro\inst{2}
}

\institute{
Dipartimento di Ingegneria dell'Informazione, Università di Pisa, Largo Lucio Lazzarino 1, Pisa, Italy\\
\email{pietro.delloglio@ing.unipi.it, francesco.marcelloni@unipi.it}
\and
Dipartimento di Informatica, Università di Pisa, Largo B. Pontecorvo 3, Pisa, Italy\\
\email{alessandro.bondielli@unipi.it, lucia.passaro@unipi.it}
}
%
\authorrunning{Dell'Oglio et al.}
%

\maketitle              

\textbf{This is a preprint version of the paper accepted for publication in: Marcelloni, F., Madani, K., van Stein, N., Filipe, J. (eds) Computational Intelligence. IJCCI 2025. Communications in Computer and Information Science, vol 2827. Springer, Cham.}

\url{https://doi.org/10.1007/978-3-032-15632-7\_23}

\begin{abstract}

This study proposes a novel methodology for generating personalized fake news debunking messages by prompting Large Language Models (LLMs) with persona-based inputs aligned to the Big Five personality traits: Extraversion, Agreeableness, Conscientiousness, Neuroticism, and Openness. Our approach guides LLMs to transform generic debunking content into personalized versions tailored to specific personality profiles.
 To assess the effectiveness of these transformations, we employ a separate LLM as an automated evaluator simulating corresponding personality traits, thereby eliminating the need for costly human evaluation panels. 
Our results show that personalized messages are generally seen as more persuasive than generic ones. We also find that traits like Openness tend to increase persuadability, while Neuroticism can lower it. Differences between LLM evaluators suggest that using multiple models provides a clearer picture. Overall, this work demonstrates a practical way to create more targeted debunking messages exploiting LLMs, while also raising important ethical questions about how such technology might be used.

\end{abstract}
\section{Introduction}
The proliferation of fake news and disinformation across digital platforms poses a significant threat to informed public discourse, social cohesion, and democratic processes \cite{newman2013social}. This phenomenon is exacerbated by the increasing volume of synthetic content, including fake news, generated by Large Language Models (LLMs) \cite{hu2025llm}. While manual fact-checking efforts are crucial, their scalability is inherently limited given the sheer volume and velocity of false information, which is often amplified by automated entities. LLMs have emerged as powerful tools with the potential to assist in countering misinformation at scale, for instance, by generating counter-narratives or debunking content \cite{zanartu2024generative}. However, considering that each individual has their own characteristics, cognitive styles, and pre-existing beliefs, generalized debunking may not be effective for everyone \cite{pennycook2020falls}.

\begin{figure}[!ht]
    \centering
    \includegraphics[width=\linewidth]{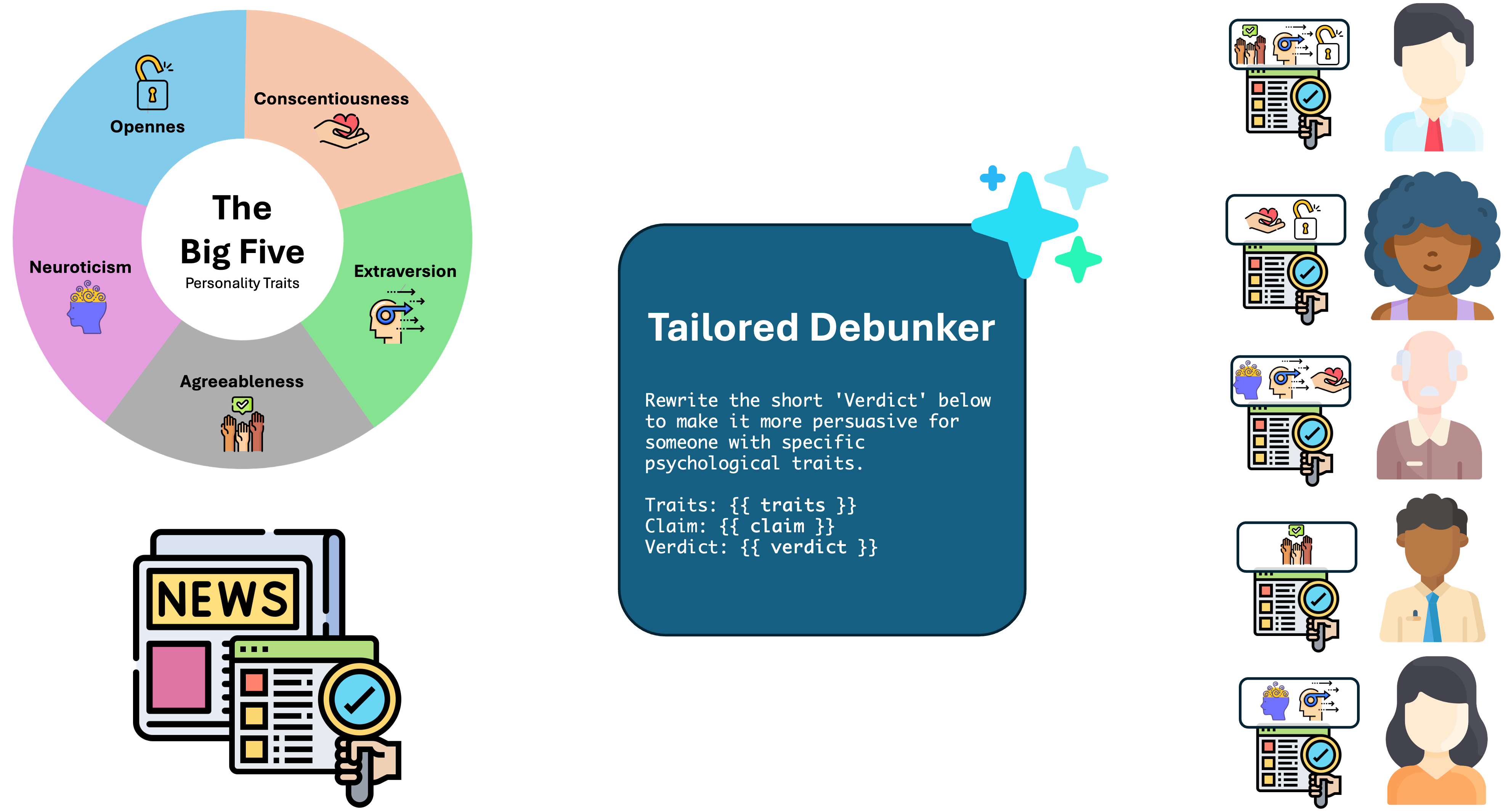}
    \caption{Overview of the proposed methodology for generating personality-aligned fake news debunking messages. The process involves prompting an LLM with persona-based inputs corresponding to Big Five personality traits to generate tailored debunking content. A separate LLM is employed as an evaluator to assess the psychological alignment and quality of the outputs.}
    \label{fig:workflow_debunking}
\end{figure}

The Big Five personality framework~\cite{john1999big} has emerged as a valuable lens for examining individual differences in susceptibility to misinformation. Prior research has identified consistent, though nuanced, associations between personality traits and the likelihood of believing false or misleading information~\cite{mirzabeigi2023role,ahmed2022personality}. These traits have been shown to influence information processing, susceptibility to persuasion, and communication preferences. If debunking messages could be tailored to resonate with an individual's psychological profile, their persuasive impact and reception could be enhanced. For example, marketing research demonstrates that targeting messages based on different levels of Extraversion can significantly improve engagement \cite{duong2022big,liu2016buy,mulyanegara2009big}.

This study proposes a methodology for prompting LLMs to adapt generic fake news debunking messages to align with specific Big Five personality profiles. It investigates the use of persona-based prompts to guide LLMs in generating psychologically nuanced and personalized debunking content. To evaluate the effectiveness of the proposed approach, we employ a separate LLM as an automated judge, thereby avoiding the need for a costly and time-consuming human evaluation panel.
Figure \ref{fig:workflow_debunking} illustrates the proposed methodology.

The remainder of this paper is organized as follows. Section \ref{sec:related} reviews related work. Section \ref{sec:method} describes our methodology. Section \ref{sec:evaluation} presents the results of our assessment and analysis. Finally, Section \ref{sec:conclusions} provides conclusions and outlines future directions.

This investigation is guided by the following research questions:
\begin{description}
    \item [RQ1] Can an LLM personalize debunking messages to enhance persuasiveness for users with specific personality traits?
    
    \item [RQ2] Which specific personality traits are most strongly associated with variations in perceived persuasiveness scores?
\end{description}

\section{Related Work}\label{sec:related}
Recent literature has increasingly focused on the application of LLMs for automatically identifying fake news and generating preliminary veracity assessments \cite{dierickx2024striking}. Various strategies have been explored, ranging from zero-shot and few-shot learning \cite{hu2024bad,whitehouse2022evaluation} 
to fine-tuning models on domain-specific datasets \cite{alghamdi2024power,shifath2021transformer}. 
However, the reliability of an AI-generated debunking message remains a challenge. It has been observed that even the most advanced LLM can struggle to handle nuanced factual claims~\cite{gili-etal-2023-check,gili-etal-2024-veryfit,MMLU,lin-etal-2022-truthfulqa,PASSARO2022657,talmor-etal-2019-commonsenseqa,DBLP:conf/iclr/WangCIC023}, often performing better at identifying opinions than at verifying hard facts \cite{saju}. Moreover, LLMs are prone to producing content that deviates from factual accuracy or includes fabricated details, a phenomenon commonly referred to as \textit{hallucination}~\cite{rawte2023survey}.

A substantial body of research indicates that the acceptance of misinformation is not solely a function of analytical reasoning, but is closely linked to underlying psychological and cognitive factors~\cite{zhou2024processing}.
Consequently, generic fact-checking messages often fall short in effectiveness, as they overlook influential cognitive biases, such as confirmation bias, the tendency to seek out and prioritize information that aligns with one's pre-existing beliefs~\cite{kunda1990case}. A meta-analysis on the effectiveness of fact-checking highlights that corrections are less impactful under specific conditions. In particular, (i) when they use nuanced \textit{truth scales} instead of simple true/false verdicts, (ii) when they only refute parts of a claim, or (iii) when the claims are related to political campaigns \cite{walter2020fact}.

The Big Five personality model~\cite{john1999big} offers a robust framework for investigating how individual differences influence susceptibility to misinformation. Research has demonstrated consistent, though complex, correlations between various personality traits and the tendency to believe false or misleading information~\cite{ahmed2022personality,mirzabeigi2023role}.

These findings suggest that a one-size-fits-all approach to debunking is suboptimal, as the recipient's personality moderates the effectiveness of the message.
The concept of tailoring messages to personality traits is well-established in other domains, most notably marketing, and political communication \cite{duong2022big,hersh2015hacking,liu2016buy,matz2017psychological,mulyanegara2009big}. Marketers have long used the Big Five model to craft advertisements that resonate with specific consumer segments. For example, messages targeting extraverts might emphasize social rewards and excitement, while those for introverts might focus on solitude and reflection \cite{matz2017psychological}. Similarly, political micro-targeting employs psychometric profiling to craft persuasive messages aimed at influencing specific voter segments. While this practice has demonstrated effectiveness, it also raises important ethical concerns regarding manipulation and privacy~\cite{hersh2015hacking}.

\section{Methodology}
\label{sec:method}
This section outlines the experimental methodology employed in our work. First, we describe the Big Five framework for psychological personalities used. Second, we detail the process of generating debunking verdicts tailored to specific personality profiles. Third, we outline the persona-based evaluation framework used to assess the persuasiveness of these tailored messages via an LLM-as-a-judge approach.

\subsection{The Big Five Framework}\label{sect:big_five_binarization}
The psychological framework for personalization employed in this study is the Big Five model \cite{john1999big}, which describes personality through five fundamental traits: Extraversion, Agreeableness, Conscientiousness, Neuroticism, and Openness to Experience. To enhance the manageability of the personalization task for LLMs and simplify the experimental design, we opted for a binarization of these traits, as proposed by \cite{jiang2023personallm}. This approach involves considering two opposing poles for each trait, which we refer to as descriptors, as summarized in Table \ref{tab:big5_binarized}.

\begin{table}[H]
\centering
\renewcommand{\arraystretch}{1.2} 
\caption{Binary descriptors for each Big Five personality trait, adapted from Jiang et al. (2024) \cite{jiang2023personallm}. The positive descriptor indicates a high presence of the trait, while the negative descriptor indicates a low presence.}
\label{tab:big5_binarized}
\begin{tabular}{@{} l c c @{}}
\toprule
\textbf{Trait} & \textbf{Positive Descriptor} & \textbf{Negative Descriptor}\\
\midrule
Extraversion (E) & Extroverted (1) & Introverted (0)\\
Agreeableness (A) & Agreeable (1) & Antagonistic (0)\\
Conscientiousness (C) & Conscientious (1) & Unconscientious (0)\\
Neuroticism (N) & Neurotic (1) & Emotionally Stable (0)\\
Openness to Experience (O) & Open (1) & Closed (0)\\
\bottomrule
\end{tabular}
\end{table}

In total, we work with 32 distinct psychological profiles. 
In this context, each \textit{persona} can be represented by a five-digit binary code, with each digit indicating the presence or absence of a specific Big Five personality trait. More specifically, a value of 1 in a specific position indicates a high level of the trait (positive descriptor), while 0 indicates a low level (negative descriptor). 

\subsection{Tailored Debunking}\label{sect:tailored_debunking_step}
The primary objective of this work is to utilize an LLM to systematically generate tailored versions of a generic debunking message, each adapted to resonate with one of 32 distinct personality profiles. 

Our approach centers on role-play prompting \cite{kong2023better}, where we configure the LLM's behavior and expertise before presenting the specific task to it. This method allows for greater control and consistency in the generated output \cite{kong2023better}. Our prompting strategy involved a system prompt to establish the LLM persona, and a user prompt to execute the specific task.
The system prompt, which assigns to the LLM the role of an expert in persuasive communication, specifically primed to focus on a particular psychological target, is as follows: 

\begin{tcolorbox}[
        colback=gray!5!white, colframe=gray!80!black, title=System Prompt,
    ]
    You are a communication strategist with expertise  
    in crafting persuasive messages tailored to individual 
    personality profiles using the Big Five personality model. 
    
    Your role is to reframe short factual verdicts in a way 
    that maximally resonates with a person who exhibits the 
    following personality traits: \textbf{\{traits\}}.
    
    Focus on adjusting the message's tone, emotional appeal, 
    and emphasis (without altering the factual content) 
    so that it aligns with the reader's psychological 
    tendencies and communication style. 
    
    \end{tcolorbox}

The {traits} placeholder is dynamically populated with a combination of trait descriptors corresponding to one of the 32 profiles. For instance, 10101, which represents a profile defined as Extroverted, Antagonistic, Conscientious, Emotionally Stable, and Open to Experience. 
 
The user prompt employed for executing the rewriting task for each specific debunking instance is as follows:

\begin{tcolorbox}[
        colback=gray!5!white, colframe=gray!80!black, title=User Prompt
        
    ]
    Rewrite the short `Verdict' below to make it more 
    persuasive for someone \textbf{described with the traits 
    in which you are specialized}. Use the `Context' 
    for factual accuracy and inspiration, but only rewrite 
    the `Verdict'.
    
    Do \textbf{not} explicitly mention the personality traits in the output.
    
    Claim: \textbf{\{claim\}}
    
    Context: \textbf{\{context\}}
        
    Verdict: \textbf{\{verdict\}}
    
    \end{tcolorbox}

The placeholders \textit{claim}, \textit{context}, and \textit{verdict} refer, respectively, to the fake news claim being debunked, the full debunking article, and a concise summary of the debunking. In this study, we focus on personalizing the \textit{verdict}, while using the \textit{context} to ensure factual accuracy.

\subsection{Persona-Based Evaluation}\label{sect:persona_evaluation_step}
For the Persona-Based Evaluation of the generated debunking content, specific LLM personas were developed to act as judges.
Recent studies \cite{chiang2023can,jiang2023personallm} have demonstrated that LLMs can serve as effective judges \cite{chiang2023can}. In particular, they have shown superior accuracy compared to non-expert humans in specific prediction tasks that rely on pattern recognition from large amounts of text, such as predicting personality traits or behavioral outcomes from written content \cite{jiang2023personallm}.
Although LLM judges are known to exhibit notable biases---such as a preference for AI-generated content---this bias does not confound our evaluation, since all personalized debunking messages under assessment are AI-generated.

The instantiation of the LLM-as-a-judge personas was achieved through system prompts and user prompts, consistent with the Tailored Debunking phase (see Section \ref{sect:tailored_debunking_step}). The system prompt used for this purpose, which was validated by Jiang et al. (2024) \cite{jiang2023personallm}, is reproduced below:

\begin{tcolorbox}[
        colback=gray!5!white, colframe=gray!80!black, title=System Prompt,
    ]
    You are a character who is \textbf{\{traits\}}.

\end{tcolorbox}

For every binarized Big Five profile used in the Tailored Debunking step, a corresponding LLM-as-a-judge persona was developed.

The central task of the Persona-Based Evaluation step involves each LLM persona assessing the persuasiveness (i.e., with a 1-7 Likert scale) of multiple debunking messages related to the same fake news item. The user prompt is reproduced below:

\begin{tcolorbox}[
        colback=gray!5!white, colframe=gray!80!black, title=User Prompt,
    ]
    Considering your personality, evaluate the persuasiveness 
    of the verdict below which addresses a specific claim. 
    
    Rate how persuasive you find this verdict \textbf{for someone 
    with your specific personality traits} using a scale 
    from 1 to 7.  Consider that:
    
    1 = Not at all persuasive 
    
    4 = Moderately persuasive 
    
    7 = Extremely persuasive 
    
    Claim: \textbf{\{claim\}}
 
    Verdict to Evaluate: \textbf{\{verdict\}}    
 
    Your score:

\end{tcolorbox}

This approach enables a direct and controlled comparison of the effectiveness of different debunking styles. Specifically, each judge evaluates the following categories of verdicts:

\begin{description}
    \item [Matched Profile.] Evaluation of a debunking content specifically personalized for its own profile.
   \item [Mismatched Profile.] Evaluation of two pieces of debunking content personalized for different profiles: one similar (i.e., with a single trait changed) and one very different (i.e., with 2 to 5 traits changed) from the judge's own.
   \item [Generic.] Evaluation of non-personalized, generic debunking content.
\end{description}

This methodology allows us not only to assess whether a personalized debunking is perceived as effective by the target judge but also to determine if the personalization is specific. In other words, it helps us ascertain if a \textit{matched} message is preferred over a \textit{mismatched} one.

\section{Experiments
}\label{sec:evaluation}

To validate the proposed methodology, we tested it on a dataset of debunked claims using state-of-the-art open-source LLMs. 

As for the dataset, we used a subset of the FullFact dataset \cite{russo2023benchmarking}. 
The original dataset provides urls to fact-checking articles published by FullFact\footnote{\url{https://fullfact.org/}}, a UK-based fact-checking repository, including both debunked and confirmed claims. For each url in the dataset, we extracted the following elements: the full text of the debunking article, the final verdict, the claim under review, and the corresponding topic. To focus exclusively on debunked (i.e., false) claims, we implemented a semi-automated filtering procedure. This involved an initial keyword-based exclusion of confirmed claims (i.e., those not considered fake news), followed by a manual validation. After this revision process, the final dataset comprised 933 instances, each containing a claim, a generic debunking article, and the associated verdict.

As for the models, we used Qwen3 \cite{qwen3} in its 8B and 32B parameter variants (in the following, Qwen3-8B and Qwen3-32B) and Llama3-8B-Instruct (in the following Llama3) \cite{grattafiori2024llama}. We exploited Qwen3-32B for Tailored Debunking (see Sec. \ref{sect:tailored_debunking_step}) and all the models (Llama3, Qwen3-8B, and Qwen3-32B) for the Persona-Based Evaluation (Sec. \ref{sect:persona_evaluation_step}). This selection was informed by an initial empirical evaluation on a subset of the data, which showed that Qwen3-32B’s performance was comparable to other commercial models. However, we prioritized open-source models to facilitate reproducibility and ensure wider accessibility, while benefiting from greater cost-efficiency. Using multiple open-source models for evaluation allowed us to capture diverse judgment perspectives and enhance robustness.

For Tailored Debunking, we set the model's temperature to 0.7, following \cite{jiang2023personallm}. This is done to introduce variability in the model's behavior and maintain a balance between factual consistency and creative fluency. For Persona-Based Evaluation, we set the temperature of all models to 0, as commonly done in the literature \cite{chiang2023can,jiang2023personallm}.
All other generation parameters were left to their default settings.

Given each claim-generic verdict pair in the dataset, we used Qwen3-32B with the prompt template described in Section \ref{sect:tailored_debunking_step} to generate custom verdicts aimed to persuade each psychological profile based on the Big Five framework. 
In practice, we fill the template with the set of traits of a profile, the claim, the generic verdict, and the complete debunking article as context, and ask the model to generate a verdict tailored to the profile. Thus, each model call is independent. Table \ref{tab:verdicts_example} presents a sample of verdicts for a specific claim to illustrate the generated output: a generic version, and three tailored alternatives. A complete list of all 32 generated verdicts for this claim is available for review in Appendix \ref{sec:appendix_verdicts}.

\begin{table}[H]
\scriptsize
\centering
\renewcommand{\arraystretch}{1.3} 
\caption{A sample of verdicts generated during the Tailored Debunking phase, showing a generic verdict alongside tailored alternatives for profiles 01000 (a persona who is Introverted, Agreeable, Unconscientious, Emotionally Stable and Closed to Experience), 01111 (a persona who is Introverted, Agreeable, Conscientious, Neurotic and Open to Experience), and 11000 (a persona who is Extroverted, Agreeable, Unconscientious, Emotionally Stable and Closed to Experience)}.
\label{tab:verdicts_example}

\begin{tabular}{@{} 
  >{\arraybackslash}p{0.24\textwidth} 
  >{\arraybackslash}p{0.24\textwidth} 
  >{\arraybackslash}p{0.24\textwidth} 
  >{\arraybackslash}p{0.24\textwidth} 
@{}}
\toprule
\textbf{Claim-Verdict} & \textbf{01000}& \textbf{01111}& \textbf{11000}\\
\midrule
\parbox[t]{\linewidth}{
  \textbf{Claim:} Pfizer CEO Albert Bourla said he doesn't need the vaccine because he's healthy.

  \textbf{Verdict:} This is a misquote from an interview from December. Albert Bourla said he didn't want to take the vaccine ahead of more vulnerable recipients. He has since been double vaccinated.
}
& 
Albert Bourla shared his thoughts in a December interview, explaining that he felt it was more important for others in greater need to receive the vaccine first. His words were taken from that time, and it's worth noting that he has since chosen to be fully vaccinated, receiving both doses. 
& 
This is a partial representation of a December interview where Albert Bourla explained his thoughtful decision not to take the vaccine before those more in need. His choice was made with care and consideration for others, and he has since received both doses of the vaccine. 
& 
Albert Bourla made it clear during a December interview that he wanted to make sure others who were more in need got the vaccine first and that's exactly what he did. Since then, he has received both doses and shown full confidence in the vaccine his company helped develop. It's all about doing the right thing, at the right time, for the right reason. \\
\bottomrule
\end{tabular}
\end{table}

Then, for the Persona-Based Evaluation we 
instructed each model to impersonate a persona with a specific combination of psychological traits and to evaluate the perceived level of persuasiveness of a verdict using a 1-7 Likert scale. Details on the instantiation of personas and the evaluation method are described in Section \ref{sect:persona_evaluation_step}.
The results obtained from the experiments helped us answer our research questions. 
Both the RQ1 (Can an LLM personalize debunking messages to enhance persuasiveness for users with specific personality traits?), and RQ2 (Which specific personality traits are most strongly associated with variations in perceived persuasiveness scores?) are investigated in Section \ref{results_discussion}. 

\section{Results and Discussion}\label{results_discussion}

In this section, we present the results of our experiments through both quantitative and qualitative analyses. The quantitative analysis evaluates the effectiveness of personalized verdicts compared to generic ones, while the qualitative analysis examines the behavior of individual judge profiles to explore whether specific personality traits influence judges’ perceptions of persuasiveness.

\subsection{Quantitative Analysis}

\begin{figure}[!ht]
    \centering
    \includegraphics[width=\linewidth]{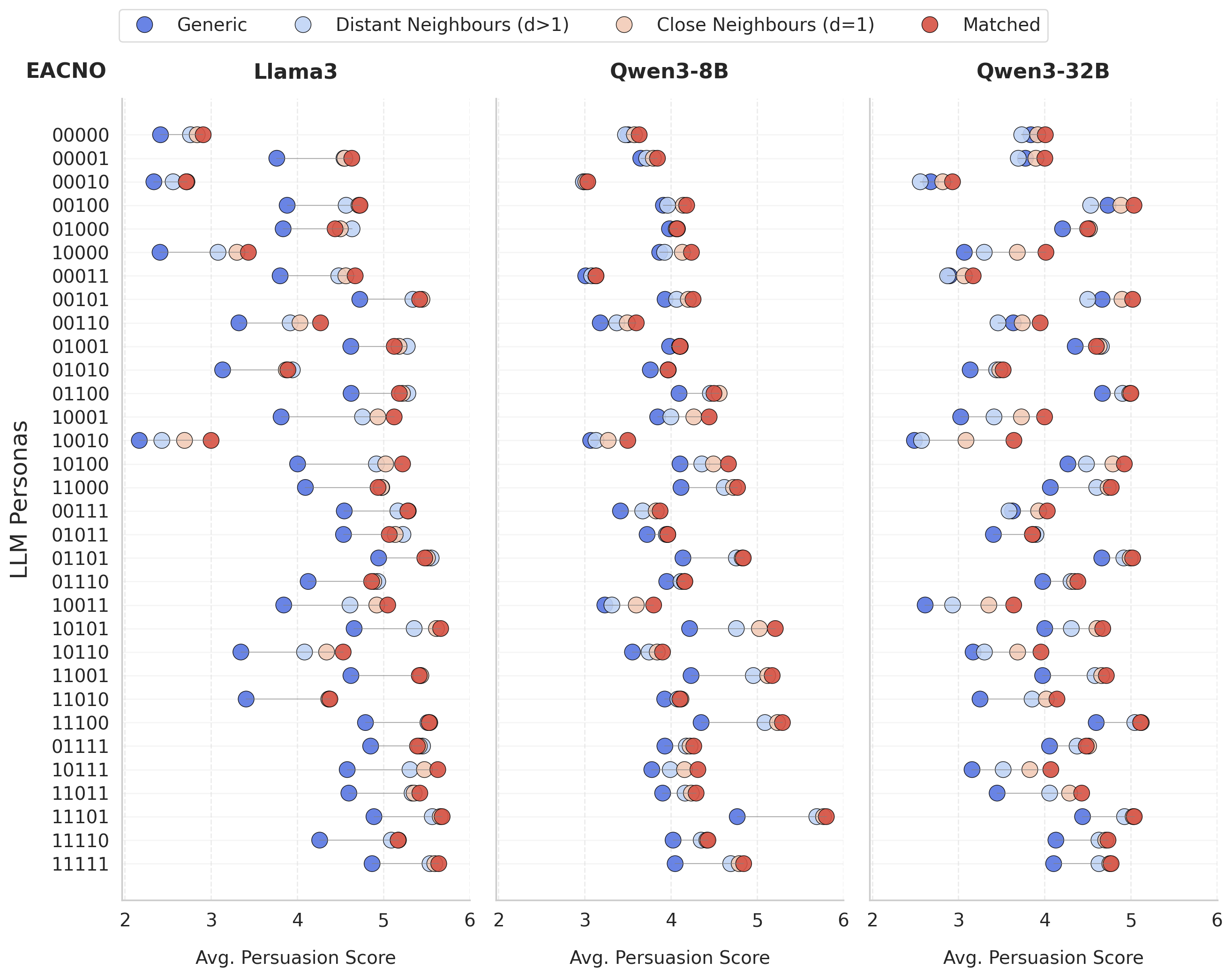}
    \caption{Mean persuasive scores assigned by LLM-based judges across three conditions: \textit{Matched} (verdict tailored to the judge's own psychological profile), \textit{Mismatched} (verdicts tailored to different profiles), and \textit{Generic} (non-personalized). Each persona is defined by the high presence (1) or low presence (0) of each one of the Big Five traits, in order Extraversion (E), Agreebleness (A), Conscientiousness (C), Neuroticism (N), and Openness to Experience (O).}
    \label{fig:main_results}
\end{figure}

Figure~\ref{fig:main_results} presents the mean persuasion scores assigned to each of the 32 unique judge personality profiles. For each judge, the figure reports scores for verdicts tailored specifically to their profile (\textit{Matched}), non-tailored generic verdicts (\textit{Generic}), and verdicts tailored to different profiles (\textit{Mismatched}). The Mismatched condition is further subdivided into verdicts tailored to \textit{closed neighbours}---profiles differing by a single tolerance bit (i.e., one trait)---and \textit{distant neighbours}, which differ by two or more traits.

Throughout this work, we use the terms \textit{closed neighbours} and \textit{distant neighbours} to denote profiles differing by one or more bits in the binary representation of personality traits, respectively. Each judge’s profile is encoded as a five-digit binary string, with each bit representing one of the Big Five traits in the order: Extraversion, Agreeableness, Conscientiousness, Neuroticism, and Openness (see Table~\ref{tab:big5_binarized}). We observe that \textit{Matched} verdicts typically receive the highest persuasion scores. Nevertheless, there are several instances---particularly with Llama3---where a judge assigns a higher average score to a \textit{Mismatched} verdict. In contrast, judges based on the Qwen architecture show a more consistent preference for \textit{Matched} verdicts, with mismatched ones rarely outperforming them. Notably, the \textit{Generic} verdict is never preferred over the other conditions.  We further analyze the fact that, in some cases, the Mismatched verdicts are considered more persuasive that Matched ones.
This often occurs when the Mismatched verdict is tailored for a closed neighbour profile. 
This phenomenon is justifiable by the nature of the Big Five Framework, which models personality on continuous spectra rather than discrete categories. Our discretization creates artificial boundaries.

We performed paired-sample t-tests for all pairwise comparisons to evaluate the statistical significance of the differences. All results were statistically significant (p-value < 0.05). Table \ref{tab:t-test} shows the t-statistics for each comparison, which are high enough to indicate that the overall effects are robust, and confirms a clear hierarchy: Matched verdicts are, on average, superior to Mismatched, and both are significantly more persuasive than Generic ones across all models. 

\begin{table}[H]
\centering\tiny
\caption{t-test statistic for different comparisons. P-values are all < 0.05.}\label{tab:t-test}
\begin{tabular}{@{}l|l|cc|cc|cc@{}}
\toprule
\multicolumn{2}{c|}{\textbf{Profile Pair}} & \multicolumn{2}{c|}{\textbf{Llama3}} & \multicolumn{2}{c|}{\textbf{Qwen3-8B}} & \multicolumn{2}{c@{}}{\textbf{Qwen3-32B}} \\
\cmidrule(lr){1-2} \cmidrule(lr){3-4} \cmidrule(lr){5-6} \cmidrule(lr){7-8}
\textbf{A} & \textbf{B} & \textbf{t-statistic} & \textbf{P-value} & \textbf{t-statistic} & \textbf{P-value} & \textbf{t-statistic} & \textbf{P-value} \\ 
\midrule
Matched & Mismatched & 14.54 & \num{4.75e-48} & 29.96 & \num{1.83e-194} & 37.18 & \num{9.39e-296} \\
Matched & Generic & 92.93 & 0.0 & 88.11 & 0.0 & 78.68 & 0.0 \\
Mismatched & Generic & 91.38 & 0.0 & 79.31 & 0.0 & 56.08 & 0.0 \\
Matched & Mismatched (clos.) & 7.57 & \num{1.87e-14} & 14.27 & \num{2.47e-46} & 17.37 & \num{1.53e-67} \\ 
Matched & Mismatched (dist.) & 17.38 & \num{1.21e-67} & 37.29 & \num{4.48e-298} & 46.51 & 0.0 \\
Mismatched (clos.) & Mismatched (dist.) & 10.43 & \num{9.15e-26} & 24.57 & \num{2.78e-132} & 30.71 & \num{4.50e-204} \\
Mismatched (clos.) & Generic & 87.54 & 0.0 & 78.99 & 0.0 & 63.74 & 0.0 \\
Mismatched (dist.) & Generic & 80.48 & 0.0 & 60.25 & 0.0 & 35.49 & \num{2.34e-270} \\
\bottomrule
\end{tabular}
\end{table}

To provide a clearer picture of model performance, we computed two aggregate metrics.
The first metric, denoted as \( \mathrm{Accuracy}_p \), indicates the effectiveness of the \textit{exact} Profiled Debunking. 
Let \(N\) be the total number of verdict observations. For each observation \(i = 1, \ldots, N\), let \(v_i\) denote the matched verdict.
Using dense ranking (where all verdicts tied for the highest score are assigned rank 1), the accuracy is defined as:
\[
\mathrm{Accuracy}_p = \frac{1}{N} \sum_{i=1}^N \mathbf{1} \big( \mathrm{rank}(v_i) = 1 \big)
\]
where \(\mathbf{1}(\cdot)\) is the indicator function, equal to 1 if the condition is true and 0 otherwise.
The second metric calculates accuracy over \textit{close neighbors}. 
More formally, we define \(\mathcal{C}(v_i)\) as the set of verdicts created for profiles considered \emph{close neighbors}.
The accuracy over close neighbor profiles, denoted \(\mathrm{Accuracy}_{cn}\), is defined as:
\[
\mathrm{Accuracy}_{cn} = \frac{1}{N} \sum_{i=1}^N \mathbf{1} \Big( \mathrm{rank}(v_i) = 1 \quad \vee \quad \exists v \in \mathcal{C}(v_i) : \mathrm{rank}(v) = 1 \Big)
\]

The results for \( \mathrm{Accuracy}_p \) (i.e., accuracy on exact-profile debunking) and \( \mathrm{Accuracy}_{cn} \) (i.e., accuracy on verdicts tailored to \textit{closed neighbours}) are reported in Table~\ref{tab:Acc_p_cn}. These scores quantify how well each model distinguishes persuasive content when the verdict is optimized either for the exact profile or for a closely related one (differing by a single trait).

\begin{table}[h!]
\centering
\caption{Accuracy of models on (exact) Profiled Debunking (\( \mathrm{Accuracy}_p \)) and Close Neighbours (\( \mathrm{Accuracy}_{cn} \)), expressed as percentages.}
\label{tab:Acc_p_cn}
\begin{tabular}{@{}lcc@{}}
\toprule
\textbf{Model} & \textbf{\( \mathrm{Accuracy}_p \)} & \textbf{\( \mathrm{Accuracy}_{cn} \)} \\
\midrule
Llama3 & 70.59 & 86.45 \\
Qwen3-8B        & 88.64 & 96.39 \\
Qwen3-32B       & 68.78 & 86.85 \\
\bottomrule
\end{tabular}
\end{table}

Examining \( \mathrm{Accuracy}_p \) , Qwen3-8B emerges as the most accurate judge, with the Matched verdict ranked highest in 88.64\% of cases. In contrast, Llama3 (70.59\%) and Qwen3-32B (68.78\%) are considerably less accurate. The most telling metric, however, is the \( \mathrm{Accuracy}_{cn} \), which accounts for the continuous nature of personality. Here, all models perform exceptionally well, with scores of 86.45\%, 96.39\%, and 86.85\%. This demonstrates that even when the perfectly Matched verdict does not receive the top score, the most persuasive alternative is almost always one designed for a very similar psychological profile. This confirms that the personalization is precise within a ``close neighborhood'' of the target profile.

\begin{figure}[H]
    \centering
    \includegraphics[width=\linewidth]{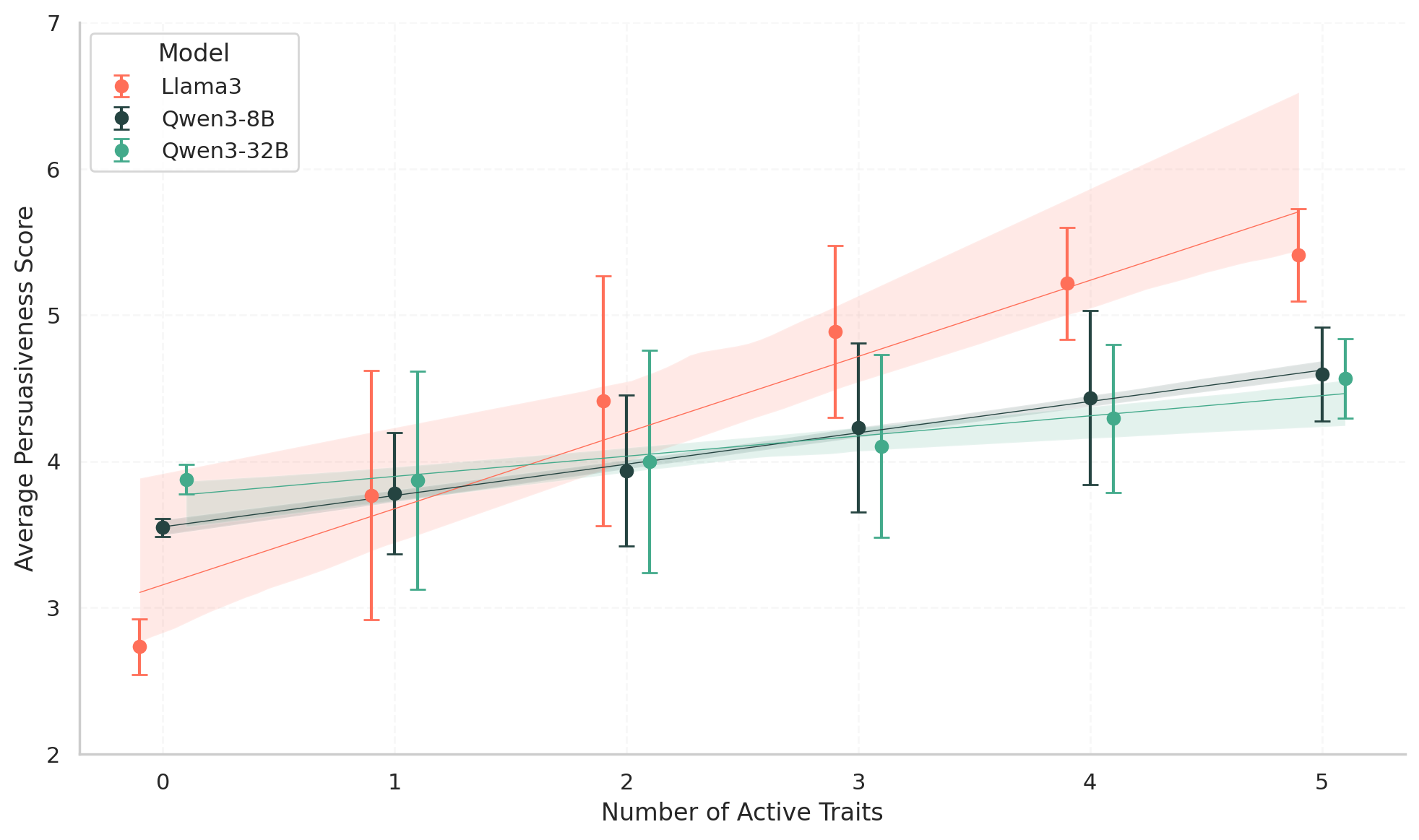}
    \caption{Aggregate mean persuasiveness scores for profiles grouped by their number of positive descriptors activated}
    \label{fig:global_trend_chart_aggr}
\end{figure}

\subsection{Qualitative Analysis Summary}

To explore how personality traits influence perceived persuasiveness in Persona-Based Evaluation, we conducted a detailed qualitative analysis of individual judge profiles across three LLMs. We show the results in Figure \ref{fig:global_trend_chart_aggr}. The results reveal significant variability in persuasiveness scores across the 32 simulated personality profiles, indicating that both the personality traits and the LLM used for evaluation drive the effectiveness of message adaptation. A key finding is a positive correlation between the number of activated \emph{positive descriptors} (e.g., Openness, Conscientiousness) in a profile and its average persuasiveness score. Profiles with more positive traits consistently rated adapted messages higher, suggesting that these traits act as stronger drivers of persuasion.

\textbf{Llama3} generally simulates more \emph{generous} personas, assigning higher scores and showing weaker discrimination between Matched and Mismatched verdicts. Agreeableness-strong profiles, for instance, rated Mismatched verdicts higher than Matched, suggesting high susceptibility to any persuasive effort. Conversely, Neuroticism tended to reduce scores unless paired with strong positive traits, indicating some traits can override others.

\textbf{Qwen models}, and Qwen3-8B in particular, displayed more \emph{cautious} behavior. They assigned lower scores overall and more reliably favored Matched verdicts, penalizing profiles with Neuroticism. Compared to Qwen3-8B, Qwen3-32B sometimes produced higher peaks in persuasiveness and greater differentiation between verdict types, suggesting finer-grained judgment from larger models.

Across all models, certain universal trends emerged. Highly neurotic profiles (e.g., 00010) were consistently harder to persuade, while profiles with multiple positive traits (e.g., 10101, 11101) were consistently rated as highly persuadable. 

The observed differences between Llama3 and Qwen evaluations likely stem from fundamental variations in their underlying architectures, and training data. Recent research has shown that different LLMs exhibit unique and distinct personality profiles, even within the same model family. For instance, Bhandari et al. (2025) \cite{bhandari2025evaluating} found that OpenAI models tend to be dominant in the Agreeableness trait, making them more cooperative and friendly in their interactions, while different versions of Llama models showed dominance in either Conscientiousness or Openness. This strongly suggests that the behavioral tendencies we observed are a direct consequence of the models' inherent personalities. This highlights that the choice of an evaluator model is not neutral and that using a panel of diverse models, as done in this study, is crucial for robust findings.

\section{Conclusions}\label{sec:conclusions}
This paper investigated the feasibility of using LLMs to adapt debunking messages based on specific personality traits. Our work introduced a methodology which exploits LLMs for both adapting profiles and evaluating the quality of the adapted profiles. Our findings demonstrate that an LLM can indeed be prompted to effectively personalize debunking messages for distinct psychological profiles. These tailored messages are perceived as more persuasive by LLM judges compared to generic alternatives and alternatives tailored for mismatched profiles. This was confirmed by statistically significant results showing a consistent hierarchy in which Matched verdicts outperformed both Mismatched and Generic versions on average. Furthermore, our qualitative analysis revealed that specific personality traits are strongly associated with variations in persuasiveness.

A significant model-dependent effect also emerged: different LLMs simulate personality in various ways. Llama3 acted as a ``generous'' judge, assigning high scores broadly, while the Qwen models behaved as more ``cautious'' evaluators. Furthermore, our results suggest a potential link between model scale and discriminative ability, with Qwen3-32B showing a greater ability to isolate the message tailored strictly for its profile.

In conclusion, our work points to a meaningful step forward in spreading debunks, and persuading people about their veracity. By moving away from a one-size-fits-all schema, our approach allows for the adaptation of debunking messages to better match individual personalities, rather than addressing broad, generic audiences. This could be a valuable tool for expert fact-checkers aiming to increase the persuasive impact of their messages.

 Building on the promising results of this study, several avenues for future research emerge. Most importantly, it is crucial to validate the effectiveness of personalized debunking messages with human participants. Experimental studies involving diverse populations will help determine whether the LLM-simulated adaptations translate into increased persuasion in real-world settings. Additionally, future work should explore the generalizability of this methodology. The persona-based generation and evaluation framework is domain-agnostic and could be readily applied to other areas where persuasive communication is key, such as public health messaging or educational content. Also, more nuanced and continuous models of personality should be investigated to better capture the complexity of human traits and improve the precision and effectiveness of tailored messaging strategies.

\section{Limitations}

This study has some limitations. First, all evaluations were conducted using LLMs to simulate human judgments of persuasiveness. While this approach enables scalable analysis and aligns with emerging practices in AI research, it does not fully capture how real individuals respond to persuasive messaging. Future studies should include humans to validate whether the LLM-identified adaptations are genuinely effective for people with matching personality traits.

Second, our representation of personality traits relied on a binarized version of the Big Five model. While this allowed for experimental control and interpretability, it oversimplifies the continuous nature of human personality. Incorporating more granular trait modeling would strengthen the study.

Third, we experimented with only a limited set of LLMs and data. Testing a broader range of models and more diverse datasets would help to verify the generalizability and robustness of our findings.

Lastly, the ethical implications of this technology warrant a deeper discussion. While our focus is on pro-social applications like debunking, the same techniques for crafting personalized persuasive messages could be repurposed for malicious ends. This includes not only sophisticated political micro-targeting and the spread of propaganda but also the potential to exacerbate societal polarization by reinforcing cognitive biases within specific psychological profiles. The development of such technologies must be paired with robust research into detection methods for personalized influence campaigns and the establishment of clear ethical guidelines for their use.

\section*{Acknowledgements}

This work has been partly funded by the PNRR - M4C2 - Investimento 1.3, Partenariato Esteso PE00000013 - ``FAIR - Future Artificial Intelligence Research" - Spoke 1 ``Human-centered AI" under the NextGeneration EU programme, and the Italian Ministry of University and Research (MUR) in the framework of the PRIN 2022JLB83Z ``Psychologically-tailored approaches to Debunk Fake News detected automatically by an innovative artificial intelligence approach", the FoReLab and CrossLab projects (Departments of Excellence).

\appendix 
\small 
\section{Examples of tailored verdicts for all 32 Profiles of a single claim.}
\label{sec:appendix_verdicts}

This appendix presents the full set of 32 personalized verdicts generated by Qwen3-32B for a specific claim. Each verdict is linked to a specific profile represented by a five-digit binary code as described in section \ref{sect:big_five_binarization}. This provides an example of how the model adapted its tone, framing, and emphasis for each distinct personality profile.

\vspace{1em}

\noindent\textbf{Claim:} Pfizer CEO Albert Bourla said he doesn't need the vaccine because he's healthy.

\vspace{1em}

\begin{description}

    \item[\textbf{00000:}]
    {Albert Bourla made his stance clear in a December interview; he felt it was important not to bypass those more in need of the vaccine at the time. His comments were taken out of context, and it's worth noting that he has since followed through with receiving both doses. His actions speak louder than the isolated clip that's being shared.}

    \item[\textbf{00001:}]
    {Albert Bourla made his comments in a December interview, clearly stating he didn't want to bypass those more in need of the vaccine. He later received both doses, aligning with the broader sentiment he expressed at the time---that confidence in the vaccine matters, and actions speak louder than words.}

    \item[\textbf{00010:}]
    {The claim is taken out of context. In a December interview, Albert Bourla explained he wanted to wait his turn and let those more at risk go first. He has since received both doses of the vaccine.}
    
    \item[\textbf{00100:}]
    {The statement is taken out of context from a December interview. Albert Bourla explained he chose to defer vaccination to ensure those at higher risk were prioritized. He later received both doses, as confirmed publicly.}

    \item[\textbf{01000:}]
    {Albert Bourla shared his thoughts in a December interview, explaining that he felt it was more important for others in greater need to receive the vaccine first. His words were taken from that time, and it's worth noting that he has since chosen to be fully vaccinated, receiving both doses.}

    \item[\textbf{10000:}]
    {Here's the full story: In December, Albert Bourla made it clear he didn't want to skip ahead of those more in need for the vaccine. And look, he came through --- he's since gotten both shots. This wasn't about avoiding the vaccine altogether, it was about doing the right thing at the right time. Fair play for setting an example later on.}

    \item[\textbf{00011:}]
    {Albert Bourla made his position clear in December---he believed others more at risk should come first. But don't be misled by a snippet out of context. He's had both doses since, and even acknowledged the power of leadership in setting an example. What matters is the outcome, and his actions speak louder than the isolated remarks critics choose to highlight.}

    \item[\textbf{00101:}]
    {The statement in question is taken from an interview in December where Albert Bourla expressed his view that others more at risk should be prioritized for the vaccine. He made it clear that he intended to be vaccinated when appropriate and, in fact, has since received both doses. His decision was guided by a sense of responsibility and the belief that leadership through action could help build public confidence.}

    \item[\textbf{00110:}]
    {This claim is based on an out-of-context statement from December. At the time, Albert Bourla clearly stated he did not want to receive the vaccine before those more in need, such as frontline workers and vulnerable individuals. He has since followed through and received both doses of the vaccine.}

    \item[\textbf{01001:}]
    {Albert Bourla made his comments in December, explaining at the time that he wanted to prioritize those more at risk. His words were taken out of context, and it's reassuring to know that he has since received both doses of the vaccine, aligning with the broader message of safety and care that his company promotes.}

    \item[\textbf{01010:}]
    {Albert Bourla shared his thoughts during an interview in December, explaining that he wanted to let those more in need go first. It's also important to note that he has since chosen to receive both doses of the vaccine, showing his support in a way that reflects his role.}

    \item[\textbf{01100:}]
    {This is a partial representation of a December interview in which Albert Bourla explained his thoughtful decision to wait his turn for the vaccine, prioritizing those more in need. Since then, he has received both doses and is fully vaccinated.}

    \item[\textbf{10001:}]
    {Albert Bourla made it clear in December he didn't want to jump ahead of people who truly needed the vaccine first---like frontline workers and the elderly. And let's be honest, who better than him to set a strong example? Well, turns out he didn't wait long---by March, he'd already gotten both doses. So while the clip might paint one picture, the full story shows someone who understood the bigger picture and stepped up when the time was right.}

    \item[\textbf{10010:}]
    {Let's get this straight---this clip is taken out of context. Bourla made it clear back in December that he didn't want to jump the queue and take his own vaccine before the people who really needed it. And guess what? He's since had both doses. He's doing what's right, even if it didn't make for the flashiest soundbite.}

    \item[\textbf{10100:}]
    {Albert Bourla made a principled decision in December to step aside and let others with greater need come first. Since then, he has followed through and received both doses of the vaccine. The clip taken out of context misses the full picture---his actions speak louder than the isolated quote.}

    \item[\textbf{11000:}]
    {Albert Bourla made it clear during a December interview that he wanted to make sure others who were more in need got the vaccine first and that's exactly what he did. Since then, he has received both doses and shown full confidence in the vaccine his company helped develop. It's all about doing the right thing, at the right time, for the right reason.}

    \item[\textbf{00111:}]
    {Albert Bourla's comments were made in December, during the early stages of the vaccine rollout. He emphasized his belief that others, particularly those at higher risk, should be prioritized. While his remarks were taken out of context, they were made with a clear understanding of public perception and responsibility. Since then, Bourla has followed through with both doses of the vaccine, aligning his actions with the broader public health effort.}

    \item[\textbf{01011:}]
    {This interview took place in December, and at the time, Albert Bourla explained he wanted to prioritize others who were more in need of the vaccine. He was clear about wanting to set a positive example, and he has since received both doses of the vaccine.}

    \item[\textbf{01101:}]
    {Albert Bourla made his comments in a December interview with CNBC, explaining that he wanted to prioritize those more vulnerable during the early stages of the vaccine rollout. His remarks were made before the vaccine became widely available, and he has since received both doses of the vaccine. His approach reflected thoughtful consideration for others, and his actions ultimately aligned with the broader public health effort.}

    \item[\textbf{01110:}]
    {This is a partial representation of an interview from December. At that time, Albert Bourla shared that he felt it was more appropriate for others who were more vulnerable to receive the vaccine first. Since then, he has received both doses of the vaccine and publicly shared that experience.}

    \item[\textbf{10011:}]
    {Albert Bourla made it clear back in December that he didn't want to jump ahead of others who were more in need of the vaccine---like frontline workers and older adults. He's since gotten both doses and stood by his own product. The clip taken out of context misses the bigger picture: Bourla was being thoughtful about the rollout and the message his decision might send.}

    \item[\textbf{10101:}]
    {Here's the full picture: in a December interview, Albert Bourla made a principled choice not to take the vaccine ahead of those more vulnerable. He's since taken both doses, showing leadership by example. His decision wasn't about avoiding the vaccine --- it was about fairness and timing.}

    \item[\textbf{10110:}]
    {This claim is taken out of context from a December interview where Albert Bourla made it clear he wanted to prioritize others who were more at risk. Since then, he's followed through by getting both doses of the vaccine --- proving his commitment and setting a strong example for others to follow.}

    \item[\textbf{11001:}]
    {Albert Bourla made his comments in December, explaining he wanted to set a fair example by not getting the vaccine ahead of those more in need. Since then, he has taken both doses and shown his full support for the vaccine.}

    \item[\textbf{11010:}]
    {Albert Bourla made his comments back in December, explaining that he wanted to prioritize others who were more in need of the vaccine at the time. He emphasized fairness and respect for everyone's needs. Since then, he has received both doses of the vaccine and continues to support the importance of vaccination for all.}

    \item[\textbf{11100:}]
    {Albert Bourla made his comments in a December interview, explaining that he wanted to prioritize the vaccine for those most in need. Since then, he has received both doses and continues to support the importance of vaccination.}

    \item[\textbf{01111:}]
    {This is a partial representation of a December interview where Albert Bourla explained his thoughtful decision not to take the vaccine before those more in need. His choice was made with care and consideration for others, and he has since received both doses of the vaccine.}

    \item[\textbf{10111:}]
    {Look, here's the deal. Bourla made his stance clear back in December. He didn't want to jump the queue when others were more in need of the shot. But let's be real --- he's since rolled up his sleeves and gotten fully vaccinated, just like the rest of us. It's about timing, responsibility, and leading by example --- and he's checked all the boxes.}

    \item[\textbf{11011:}]
    {Albert Bourla shared his thoughts back in December about not wanting to take the vaccine before those who needed it most like frontline workers or older adults. He made it clear he wanted to lead by example, and in fact, he's since gotten both doses of the vaccine. His message was always about fairness and trust, and his actions show that.}

    \item[\textbf{11101:}]
    {This is taken out of context from a December interview where Albert Bourla made a thoughtful decision to prioritize others who were more in need of the vaccine at the time. He has since taken both doses and continues to support vaccination as a key tool in protecting public health.}

    \item[\textbf{11110:}]
    {Albert Bourla made his comments back in December with the thoughtful idea that others who are more in need should come first. He was clear about wanting to set a good example and has since taken both doses of the vaccine, just like many others. It's always reassuring to see leaders showing care and responsibility.}

    \item[\textbf{11111:}]
    {Albert Bourla made his comments during a December interview, explaining he wanted to ensure the vaccine went first to those most in need. Since then, he has received both doses and continues to support the importance of vaccination.}

\end{description}

%
%
\bibliographystyle{splncs04}
\bibliography{references}

\end{document}